\documentclass[journal]{IEEEtran}

\usepackage{color}
\usepackage{times}
\usepackage{epsfig}
\usepackage{graphicx}
\usepackage{amsmath}
\usepackage{amssymb}
\usepackage{multirow}
\usepackage{floatrow}
\usepackage{mathrsfs}
\usepackage{array}
\usepackage{subfig}
\usepackage{ifpdf}
\usepackage{algorithmic}
\usepackage{algorithm}

\usepackage[colorlinks,
            linkcolor=red,
            anchorcolor=blue,
            citecolor=green,
            urlcolor=blue
            ]{hyperref}

\ifCLASSINFOpdf
  \DeclareGraphicsExtensions{.pdf,.jpeg,.png}
    \DeclareGraphicsExtensions{.eps}
\usepackage{epstopdf}
\else
\fi

\begin{document}
\title{DDet: Dual-path Dynamic Enhancement Network for Real-World Image Super-Resolution}

\author{
Yukai Shi, Haoyu Zhong, Zhijing Yang~\IEEEmembership{Member,~IEEE}, Xiaojun Yang, Liang Lin,~\IEEEmembership{Senior Member,~IEEE}
        
        \thanks{
        Code address: \url{https://github.com/ykshi/DDet}
        }}

\markboth{}%
{Shell \MakeLowercase{\textit{et al.}}: Bare Demo of IEEEtran.cls for IEEE Journals}

\markboth{}
{Shell \MakeLowercase{\textit{et al.}}: Bare Demo of IEEEtran.cls for IEEE Journals}
\maketitle

\begin{abstract}
Different from traditional image super-resolution task, real image super-resolution(Real-SR) focus on the relationship between real-world high-resolution(HR) and low-resolution(LR) image. Most of the traditional image SR obtains the LR sample by applying a fixed down-sampling operator. Real-SR obtains the LR and HR image pair by incorporating different quality optical sensors. Generally, Real-SR has more challenges as well as broader application scenarios. Previous image SR methods fail to exhibit similar performance on Real-SR as the image data is not aligned inherently. In this article, we propose a Dual-path Dynamic Enhancement Network(DDet) for Real-SR, which addresses the cross-camera image mapping by realizing a dual-way dynamic sub-pixel weighted aggregation and refinement. Unlike conventional methods which stack up massive convolutional blocks for feature representation, we introduce a content-aware framework to study non-inherently aligned image pair in image SR issue. First, we use a content-adaptive component to exhibit the Multi-scale Dynamic Attention(MDA). Second, we incorporate a long-term skip connection with a Coupled Detail Manipulation(CDM) to perform collaborative compensation and manipulation. The above dual-path model is joint into a unified model and works collaboratively. Extensive experiments on the challenging benchmarks demonstrate the superiority of our model.
%Specifically, the proposed model achieves 1 dB improvement against state-of-the-art methods.

\end{abstract}


% \fi
\IEEEpeerreviewmaketitle

\begin{figure*}[t]
    \centering
    \includegraphics[width=0.90\textwidth]{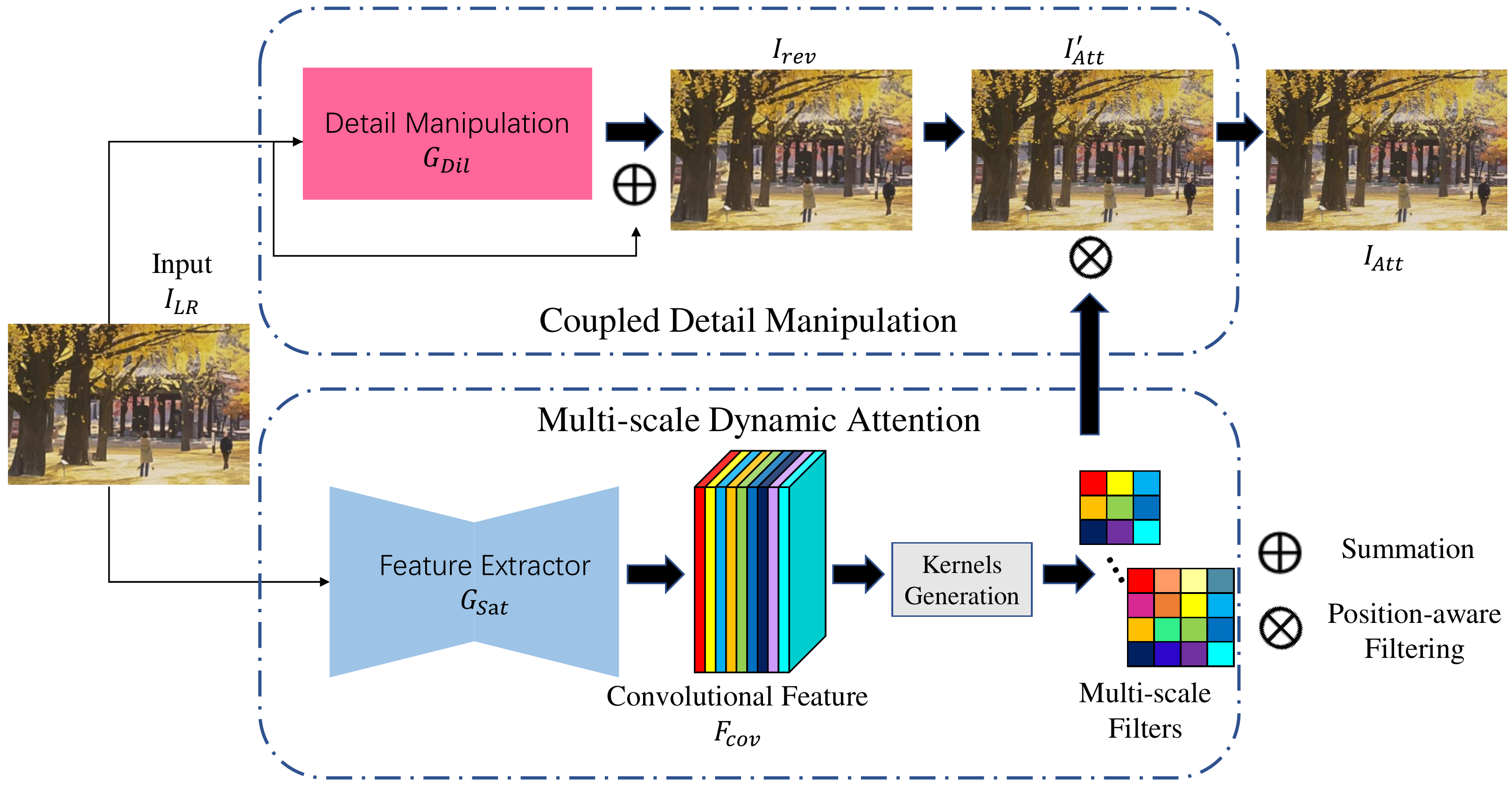}
    \caption{Illustration of our dual-path dynamic enhancement network.}
    \label{fig:intro}
\end{figure*}

\section{Introduction}
Owing to the rapid development of artificial intelligence technology, emerging applications, such as Alexa, Woogie, and Prisma, grow more prevalent than ever to change human lives. As a classical problem in computer vision, image super-resolution technology~\cite{shi2019face,ren2018nonlocal,cosmo2019multiple,chu2018novel,lu2018deep,shi2017structure,wang2018information} also achieves tremendous progress and be widely used in many mobile devices, such as mobile phones, for photo enhancement~\cite{ignatov2018pirm}. With such a light-weight algorithm, mobile devices are capable of providing a high-quality photograph and free from purchase expensive sensors. However, due to the existing image SR algorithms are restricted to handle simple down-scaling kernels(e.g., Bicubic, Bilinear), the robustness of traditional image SR models is limited when dealing with intricate data.

Example- based approaches~\cite{freeman2002example} open a new solution for image SR by a data-driven fashion and address the relationship between internal- and external- sample~\cite{tirer2019super,cheong2017deep,cao2017attention}. Yang~\emph{et al}~\cite{yang2010image} uses spare representation to build a dictionary learning algorithm between LR and HR image. Deep convolutional neural networks have also generally applied in image SR. For instance, SRCNN~\cite{dong2014learning} uses a three-layer convolution network to simulate the spare coding process. VDSR~\cite{vdsr} first adopts a long-term skip connection to enforce neural network approximate residual map(i.e., the difference between LR image and HR image), rather than image content in the value domain. Similarly, the long-term skip connection technology is widely used in the latter image SR frameworks. EDSR~\cite{edrn} employs an efficient convolutional block and long-term skip connection to realize high-quality image restoration. RCAN~\cite{rcan} also employs feature squeeze and extraction strategy and multiple lengths skip connection for advanced image SR. These methods aim at restoring the corrupted image with fixed degeneration kernel and the real-world degeneration means may complicated and diverse, which limit the restoration quality when meeting real-world photo. Also, the fixed degeneration kernel inherently makes the input and the ground-truth image has a stable relationship and therefore turns the long-term residual connection to become practical. Nevertheless, the real-world LR and HR image pair are unable to have a strict aligned relationship as the filming conditions change and absolute alignment without information loss is impractical. To this end, these methods have two drawbacks: 1) Most of the image SR methods obtain the LR input by applying fixed degradation metrics on the ground-truth image, which limited the generalization ability toward real-world cases. 2) It is potentially less optimal that forward the original corrupt input when dealing with real-world image restoration, regardless of their pixel displacement. 

Although conventional convolutional networks can extract diverse features with various inputs, the parameter is fixed during the inference process. DFN\cite{jia2016dynamic} opens up a new solution to turn convolutional features into dynamic filters for feature aggregation. Similar idea\cite{bhowmik2017training,realsr} is addressed in the image restoration task yet. KPN\cite{realsr} proposes a dynamic prediction mechanism by employing the high-dimensional feature into convolutional kernels. LP-KPN~\cite{realsr} further conducts the dynamic kernels on multi-scale feature for image restoration. As the multi-scale strategy~\cite{yang2019multilevel,huang2019pyramid,yang2019lightweight,liang2015human} prefers to conduct kind of down-samplings to obtain large receptive field, the relative local context is sufficient for restoration. Therefore, the usage of applying various size kernels is implied to aggregate local context effectively and avoid information lose.

%根据前面提及的concerns， 我们提出了dual-path dynamic enhancement network as a way of building a robust long-term skip connection and efficient multiscale information aggregation. 
This paper addresses the aforementioned issues by presenting a dual-path dynamic enhancement network. 
Our contribution is to conduct a multi-scale dynamic attention module(MDA). Compared with the traditional convolutional kernel, our spatial attention kernel is learned in a content-adaptive fashion, which is more flexible. Compared with dynamic filter pipelines~\cite{dynamic,realsr}, our model uses multiple dynamic kernels from the different receptive field without information loss, which make our method enjoys rich diversity. We also conduct a lightweight coupled detail manipulation(CDM) to relax pixel displacement in real image SR. Compared with other long-term skip connection pipeline~\cite{vdsr}, our lightweight model exhibits a supporting pixel manipulation and correction. The above two components are joint together by a dual-way fashion to demonstrate charming results on real image SR benchmark.

\section{Methodology}
\subsection{Overview}
By convention, the image super-resolution aims at restoring the high-quality image$I_{HR}$ from poor-quality input $I_{LR}$. In classical image SR issue, the $I_{LR}$ is obtained from $I_{HR}$ with a down-sampling operation:
\begin{equation}
I_{LR} = \partial \left ( I_{HR}\otimes K_{Gauss}  \right ),
\end{equation}
where $ K_{Gauss}$ is a Gaussian blur operator and $\partial$ denotes the degradation process. $\partial$ usually adopts a Bicubic down-scale algorithm in traditional image SR. Real-SR dataset~\cite{realsr} is proposed to address real-world degeneration metrics by capturing $I_{HR}$ and $I_{LR}$ with different quality optical sensors and resolution settings. The data collection manner inherently make the image pair has different resolution property.

\begin{figure*}[t]
\centering
\subfloat[Laplacian Pyramid~\cite{realsr}]{
\label{fig:kernel_1}
\begin{minipage}[t]{0.45\textwidth}
\centering
\includegraphics[width=0.99\textwidth]{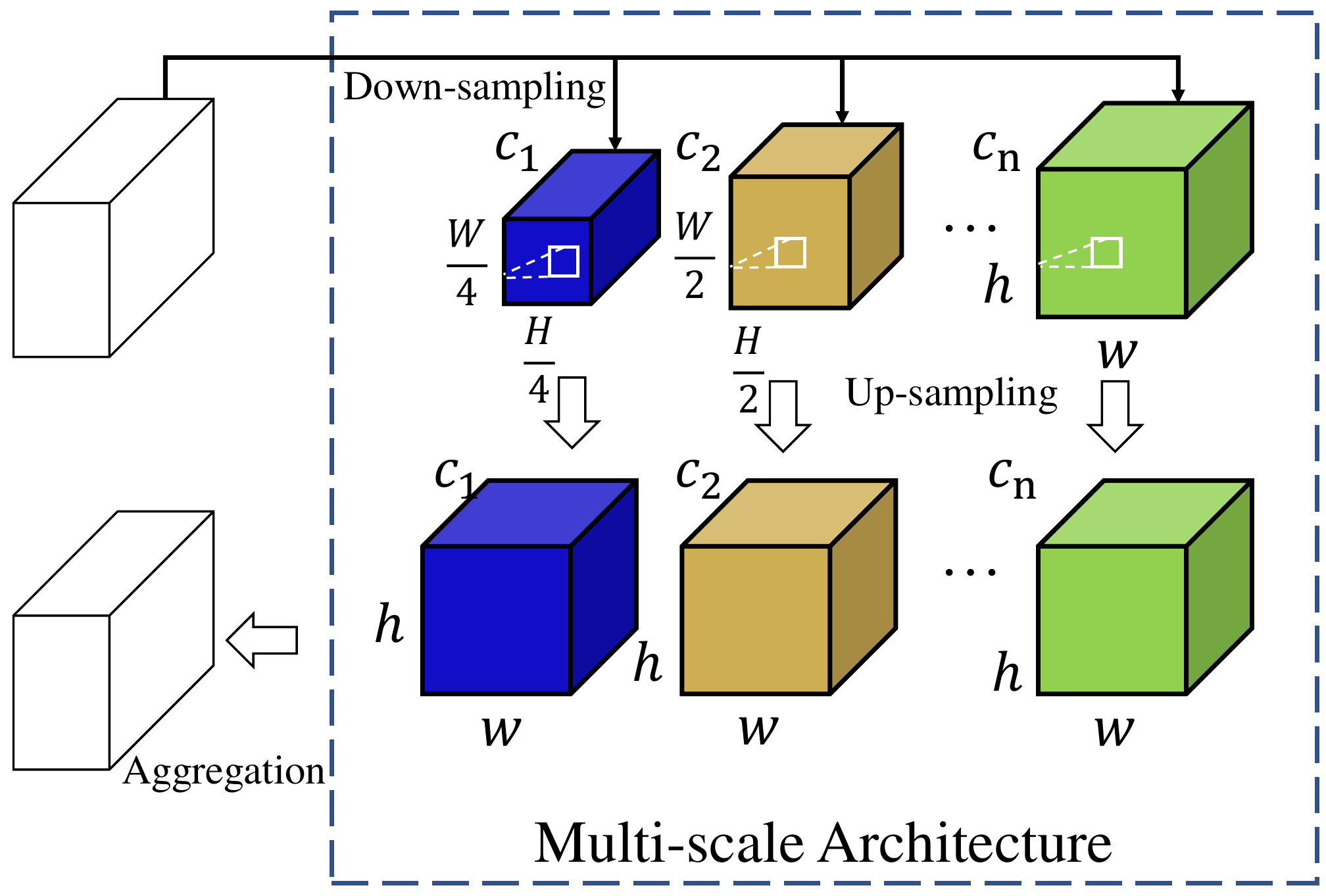}
\end{minipage}
}
\subfloat[Our]{
\label{fig:kernel_2}
\begin{minipage}[t]{0.45\textwidth}
\centering
\includegraphics[width=0.99\textwidth]{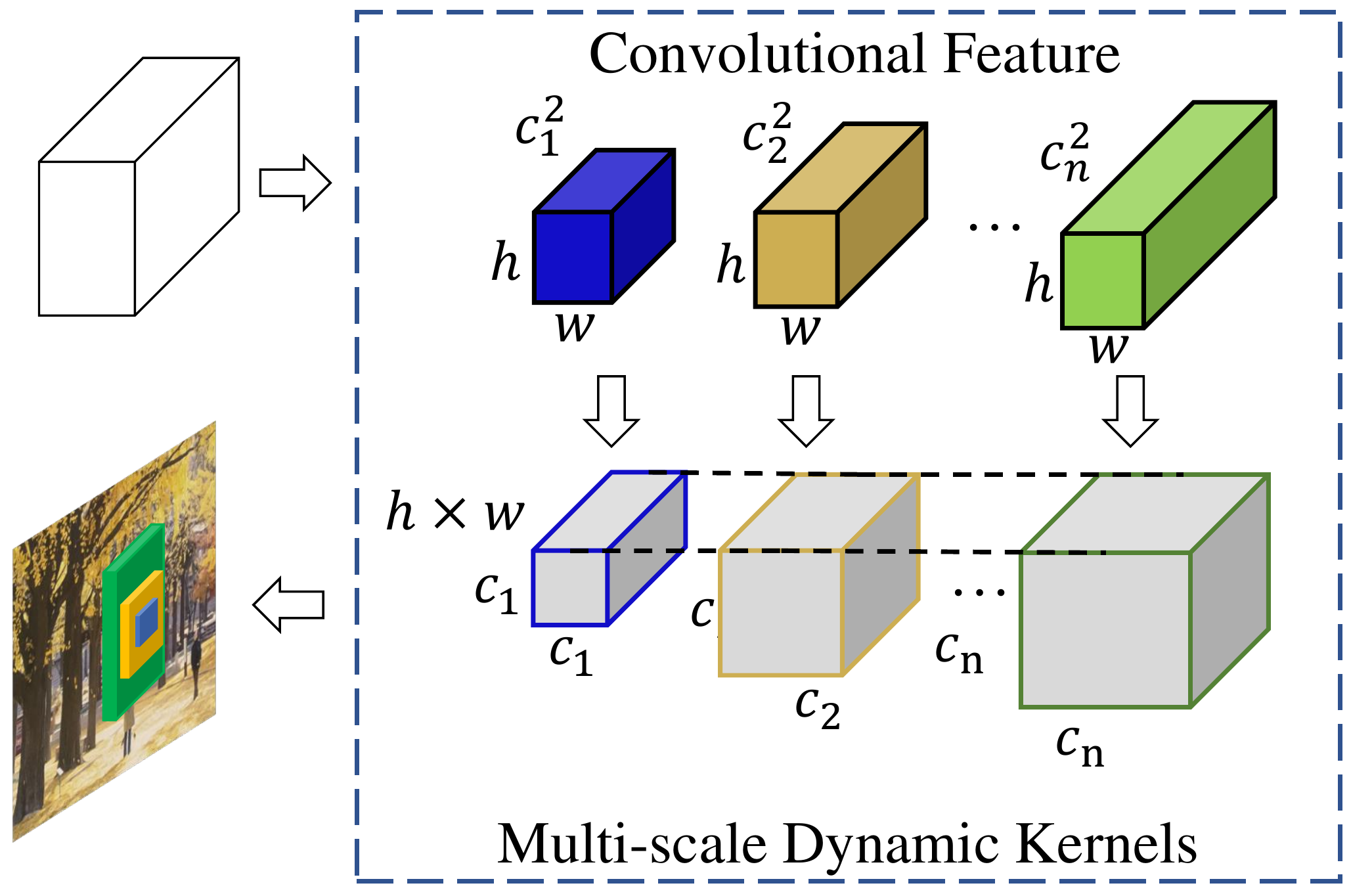}
\end{minipage}
}
\caption{Illustration of multi-scale dynamic attention mechanism. Typical multi-scale methods(e.g., (a)) tend to compress the feature into small size for the large receptive field, and then conduct convolutional feature extraction and re-scaling. Since relative local information is enough for the image SR, the proposed multi-scale dynamic attention module applies the various size of dynamic convolutional filters, which not only demonstrate a multi-scale enhancement but also free from re-scaling.
}
\label{fig:multi_scale}
\end{figure*}
DDet consists of two components(i.e., MDA and CDM). Given a low-resolution image $I_{LR}$, DDet send $I_{LR}$ into the dual-path model parallelly:
\begin{equation}
I_{Rev} = \mathbf{G}_{Dil}\left ( I_{LR} | \Theta_{Dil} \right ) + I_{LR},
\end{equation}
where the $ \Theta_{Dil}$ indicates the parameter of CDM. Since the overall reveal process based on a residual fashion, we use the $I_{LR}$ as the reference for the output of CDM. After obtaining the $I_{Rev}$, we employ a dynamic kernel to exhibit a local spatial aggregation. 
Then, we extract the $I_{LR}$ into high-dimensional feature and represent it into dynamic kernel for spatial attention:
\begin{equation}
K_{Dyn} = \mathbf{G}_{Sat} \left ( I_{LR} | \Theta_{Sat} \right ),
\end{equation}
where $K_{Dyn}$ is learned content-adaptive kernel. As the kernel is flexible and aware to image content, we conduct a content-adaptive spatial attention by:
\begin{equation}
I_{Att} = \mathbf{F_{PR}} \left (\mathbf{g} \left ( I_{Rev} | K_{Dyn} \right ) \right ),
\label{eqa:attention}
\end{equation}
where $\mathbf{g}$ indicates a convolution operator, $K_{Dyn}$ is corresponding convolutional kernels and $\mathbf{F_{PR}}$ is a post-refinement operation. The above two branch work complementary for real image SR.
\begin{figure}[t]
\centering
\subfloat[$I_{LR}$]{
\label{fig:detail_revealing1}
\begin{minipage}[t]{0.45\textwidth}
\centering
\includegraphics[width=0.99\textwidth]{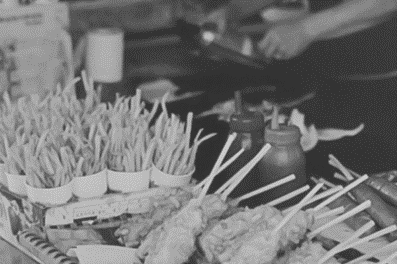}
\end{minipage}
}
\subfloat[$I_{rev}$]{
\label{fig:detail_revealing2}
\begin{minipage}[t]{0.45\textwidth}
\centering
\includegraphics[width=0.99\textwidth]{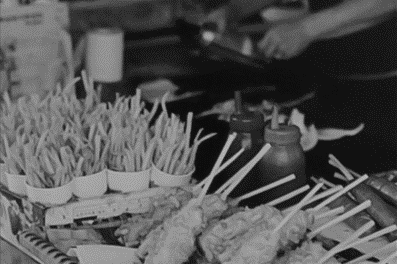}
\end{minipage}
}
\caption{Demonstration of coupled detail manipulation.}
\vspace{-3mm}
\label{fig:detail_revealing}
\end{figure}
\vspace{-3mm}
 \begin{figure*}[ht]
    \centering
    \includegraphics[width=0.91\textwidth]{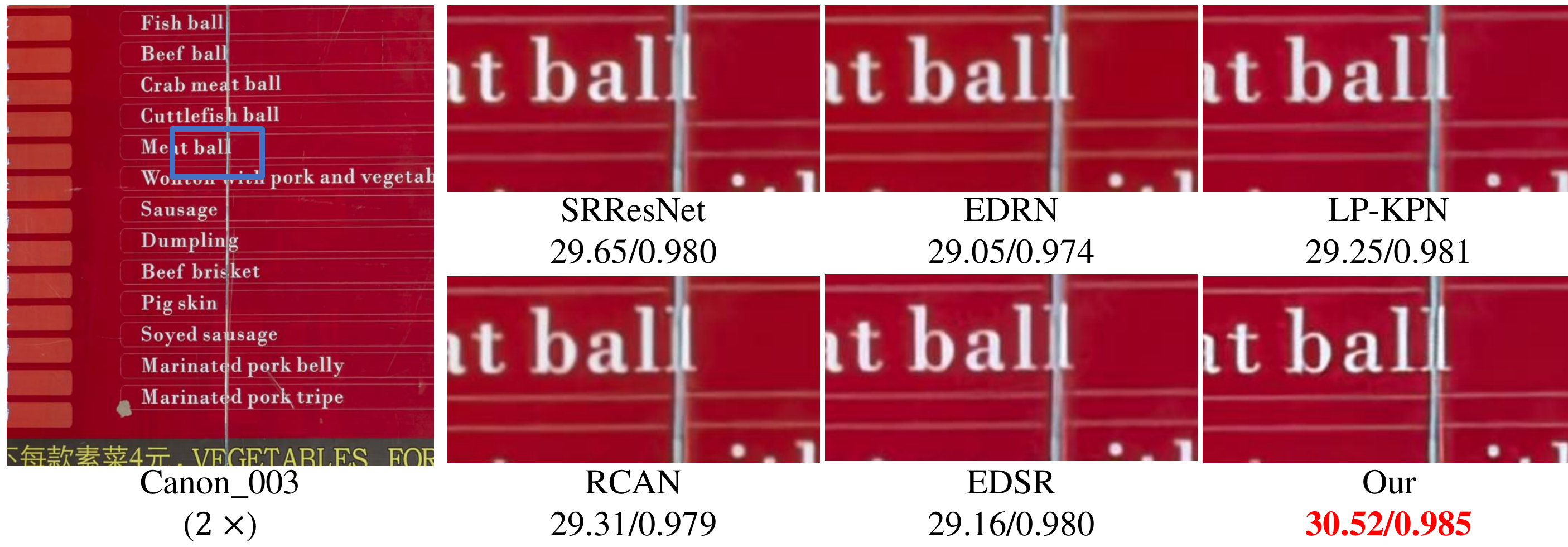}
    \vspace{-3mm}
    \caption{Visual comparison for 2 $\times$ SR. The best results in term of PSNR/SSIM are highlighted.}
    \label{fig:result}
\end{figure*}
\subsection{Multi-scale Dynamic Attention}
The MDA is used for extract valuable local information and enhance the original low-quality input into high-quality image. As Fig.~\ref{fig:intro} show, MDA consists of a feature extractor $G_{Sat}$ and the kernel generation process. Meanwhile, $G_{Sat}$ inherits the structure of auto-encoder. We first down-sampling $I_{LR}$ with scale 4$\times$:
\begin{equation}
y_{down} = \Downarrow C(I_{LR}),
\end{equation}
where $ \Downarrow $ indicates the down-sampling and $C$ means the convolutional feature extraction. We use a set of residual block for the feature representation of $y_{down}$. Each residual block consists of two convolutional layer with size of $64 \times 3 \times 3$. We use 16 residual block in $G_{Sat}$ to obtain deeper representation ${y}'_{down}$. Finally, we upscale ${y}'_{down}$ as:
\begin{equation}
\Upsilon_{Dyn} = C\Uparrow C({y}'_{down}),
\end{equation}
where $\Uparrow$ is 4$\times$ upscale operation and $\Upsilon_{Dyn}$ is final output of $G_{Sat}$. The 4$\times$ down-sampling and up-sampling are adopted for two reasons: 1) Since the low- and high- resolution images are not be aligned strictly, the 4$\times$ is able to help the network obtains larger receptive field. 2) Under the limitation of GPU memory, the 4$\times$ down-sampling and up-sampling make it possible to incorporate deeper neural architecture. According to our experiment, a deeper model with down-sampling demonstrates better results than a shallow model without down-sampling.

\textbf{Dynamic filter formulation.} We then use $\Upsilon_{Dyn}$ to formulate our feature-based dynamic convolutional filters. we reformulate the feature into kernel shape across the channel. For instance, we have a feature with a size of $N \times C \times W \times H$, which indicates batch size, channel number, feature width, and height, respectively. Meanwhile, the feature width and height own the same shape with $I_{HR}$. Hence, a single dynamic convolutional filter can be obtained with:
\begin{equation}
K^{(w,h)}_{Dyn} = \Upsilon_{Dyn}(w,h),
\label{eqa:kernel}
\end{equation}
$w$ and $h$ indicate the location. Besides, $1\leq w \leq W $ and $1\leq h \leq H$. Since we adopt a point-aware filtering strategy, each pixel in $I_{HR}$ own a individual dynamic filter. Suppose the $I_{HR}(w,h)$ is a pixel in the $I_{HR}$, we need to collect spatial information with pure local perspective by a dynamic filter $K_{Dyn}^{(w,h)}$. Therefore, $K_{Dyn}^{(w,h)}$ has corresponding location with $\Upsilon_{Dyn}(w,h)$. We obtain each element in $K_{Dyn}^{(w,h)}$  alone the channel of $\Upsilon_{Dyn}(w,h)$ as:
\begin{equation}
\begin{cases}
 & K^{(w,h)}_{Dyn}(i_1,j_1) = \Upsilon_{Dyn}(w,h,c_n), n = 1 \\ 
 & K^{(w,h)}_{Dyn}(i_1,j_2) = \Upsilon_{Dyn}(w,h,c_n), n = 2  \\ 
 & \cdots  \cdots  \\
 & K^{(w,h)}_{Dyn}(i_{n},j_{n}) = \Upsilon_{Dyn}(w,h,c_{n^2}),n^2 = C.
\end{cases}
\label{eqa:kernel_detail}
\end{equation}
Then, as illustrated in Eq.~\ref{eqa:attention}, the generated $K_{Dyn}$ is directly used on $I_{HR}$ to demonstrate a pure local attention. 
\begin{table*}[ht]
\begin{tabular}{|c|c|c|c|c|c|c|c|}
\hline
\multicolumn{2}{|l|}{\multirow{2}{*}{Methods}} & \multicolumn{2}{c|}{$\times$2} & \multicolumn{2}{c|}{$\times$3} & \multicolumn{2}{c|}{$\times$4} \\ \cline{3-8} 
\multicolumn{2}{|l|}{}                         & PSNR       & SSIM       & PSNR       & SSIM       & PSNR       & SSIM       \\ \hline \hline
\multirow{3}{*}{Traditional SR}  & SRResNet    &  33.79     & 0.918           &   30.52  &   0.858  &      29.03  &   0.821         \\ \cline{2-8} 
                                 & EDSR        & 33.57     &  0.916   &  30.13   &           0.847 &   28.77   &   0.812    \\ \cline{2-8} 
                                 & RCAN        & 33.76      &  0.917    & 30.36    &   0.861      &  28.89    &  0.820   \\ \hline \hline
\multirow{6}{*}{Real SR}         
& EDRN        &     33.72       &           0.920 &  30.21    &    0.852     &  28.79    &  0.809  \\ \cline{2-8}
                                 & KPN(K=5)    &  33.41	    &           	0.913 &  30.47	    &  0.860    &   28.80   &  	0.826     \\ \cline{2-8} 
                                  & KPN(K=7)    &  33.42	    &           	0.913 &  30.49	    &  0.861    &   28.84   &  	0.826     \\ \cline{2-8} 
                                 & KPN(K=13)   &  33.44    &           0.913	 &  30.52    &   0.863    &  	28.92	    &   0.829   \\ \cline{2-8} 
                                 & KPN(K=19)   &  33.45   &           0.914 &  30.57    & 0.864    &   	28.99   &  0.832   \\ \cline{2-8} 
                                 & LP-KPN(K=5) & 33.49     &           	0.917 &  30.60    &  0.865    & 	29.05    &  	0.834   \\ 
                                 \hline \hline
\multicolumn{2}{|c|}{Our}                      &  \textbf{ {\color{red} 34.37 }}    &   \textbf{ {\color{red} 0.932 }} &  \textbf{ {\color{red} 31.08 }}    &  \textbf{ {\color{red} 0.880 }}    &  \textbf{ {\color{red} 29.23 }}    & \textbf{ {\color{red} 0.837 }} \\ \hline
\end{tabular}
\caption{Comparison between our models and other methods on the PSNR/SSIM indexes.}
\vspace{-3mm}
\label{tab:result}
\end{table*}

\textbf{Kernels generation.} we conduct a multi-scale dynamic attention strategy, we generate multiple dynamic spatial kernels as illustrated in Eq.~\ref{eqa:kernel} and~\ref{eqa:kernel_detail}. To fully capture multi-scale information, we adopt three dynamic spatial kernels with a size of 3$\times$3, 5$\times$5 and 7$\times$7. We conduct the dynamic convolution filters work on $I_{Rev}$ directly:

\begin{equation}
{I}'_{Att} =  \sum_{n=1}^{N} \left (\mathbf{g_{n}} \left ( I_{Rev} | K^{n}_{Dyn} \right ) \right ),
\label{eqa:aggregation}
\end{equation}
$N$ denotes the kernel number, we use 3 dynamic kernels for multi-scale dynamic attention.  As shown in Fig.~\ref{fig:multi_scale}, our model adopt a multi-scale attention based on various size of dynamic convolutional filters$\left \{  K^{1}_{Dyn}, ..., K^{n}_{Dyn} \right \}$. Specifically, we incorporate several dynamic kernels on $I_{Rev}$ in parallel. The parallel results are aggregated together by summation to obtain the ${I}'_{Att}$. Since the kernel size is decided by channel number of $\Upsilon_{Dyn}$, we adopt 9, 25 and 49 channel numbers of $\Upsilon_{Dyn}$ to formulate the corresponding dynamic kernels. 

\vspace{-3mm}
\subsection{Coupled Detail Manipulation}
Suppose we have the ${I}'_{Att}$, a critical problem is the input image may have pixel displacement and disparity. Therefore, a complementary component $G_{Dil}$ is incorporated for compensation and enhancement. $G_{Dil}$ adopts a residual learning strategy for the detail revealed. Supposed we have $I_{LR}$, the residual block is incorporated for feature representation:
\begin{equation}
I_{rev} = C_{Res}(I_{LR}) + I_{LR},
\end{equation}
$C_{Res}$ means residual block. The incorporated residual block has three convolutional layers with 3 $\times$ 3 convolutional kernel, the channel number is 64, 64 and 1, respectively. Once we obtain the $I_{Rev}$, as illustrated in Eq.~\ref{eqa:aggregation}, we employ the learned $K_{Dyn}$ work on enhanced image $I_{Rev}$ for a more accurate spatial aggregation. As shown in Fig.~\ref{fig:detail_revealing}, compared with a single branch baseline, the proposed dual-way mechanism is able to realize a compensation and correction on original input, which exhibits an accurate image enhancement. Finally, we introduce a post-refinement operation by employing a convolution kernels with size of 3$\times$3 on ${I}'_{Att}$ for additional enhancement and restoration:
\begin{equation}
I_{Att} = \mathbf{F_{PR}}   \left ( {I}'_{Att},\Theta_{PR} \right )
\end{equation}

% or


\section{Experiment}
\textbf{Dataset and Baselines.} We employ real-world single image super-resolution dataset(RealSR)~\cite{realsr} to evaluate the proposed method. RealSR dataset has 3,147 images, which are collected from 559 scenes. The images are mainly captured from Canon 5D3 and Nikon D810 devices. The cross-sensor image pair are aligned with iterative pixel-wise registration and alignment. RealSR dataset contains three scale(i.e., 2$\times$, 3$\times$ and 4$\times$). Each scale has 400 image pairs for training and 100 image pair for testing. Meanwhile, the training samples are collected from 459 scenes and testing images are collected from 100 scenes. To justify the effectiveness of our model, we conduct traditional image SR approaches, including SRResNet~\cite{srgan}, RCAN~\cite{rcan} and EDSR~\cite{edsr} for comparison. Furthermore, we also employ real image SR methods, including KPN~\cite{realsr}, EDRN~\cite{edrn}, for comparison.
\begin{table}[]
\begin{tabular}{|c|c|c|c|c|}
\hline
Algorithm & Plain &  w/ PR & w/ CDM &  w/ MDA \\ \hline
PSNR      & 33.43 & 33.80     & 34.28      & 34.37      \\ \hline
\end{tabular}
\caption{Ablation study on RealSR dataset with scale $\times$2}
\label{tab:ablation}
\vspace{-1mm}
\end{table}

\textbf{Comparison.}
As Tab.~\ref{tab:result} shown, our model achieves promising results on the real image SR benchmark. Compared with LP-KPN, our model achieves 0.88 dB promotion. Compared with RCAN~\cite{rcan}, our method exhibit a 0.61 dB gain. It notes that EDSR gets into overfitting as the training sample number is limited. As shown in Fig.~\ref{fig:result}, our model restore a clear structure with less artifact, which justifies that the restoration quality improvement of our model is significant. 

\begin{table*}[ht]
\begin{tabular}{|c|c|c|c|c|}
\hline
\multicolumn{2}{|l|}{}                      & Time(s) / Frame    &   Patameter(MB) & PSNR(dB)\\ \hline
\multirow{3}{*}{Traditional SR} & SRResNet  &   2.1            & 4.9  &  30.52 \\ \cline{2-5} 
                                & EDSR      &   8.7              &  153.5  & 30.13 \\ \cline{2-5} 
                                & RCAN      &  3.61               &  54.8   & 30.36 \\ \hline
\multirow{6}{*}{Real SR}        & EDRN      &  0.29             &  131.9   & 30.21  \\ \cline{2-5} 
                                & KPN(K=5)  &   0.042         &  5.1   &  30.47 \\ \cline{2-5} 
                                & KPN(K=7)  &   0.047        &  5.2    & 30.49 \\ \cline{2-5} 
                                & KPN(K=13) &   0.065             &  5.5  &  30.52 \\ \cline{2-5} 
                                & KPN(K=19) &  0.087             &  5.9  &  30.57  \\ \cline{2-5} 
                                & LP-KPN(K=5)  &   0.138             &  5.7 &  30.60  \\ \hline
\multicolumn{2}{|c|}{Our}                   &  0.123      &   5.7  & 31.08  \\ \hline
\end{tabular}
\caption{Efficiency and parameter comparison with down-sample factor $\times$3 on RealSR dataset~\cite{realsr}. }
\label{tab:efficiency}
\end{table*}

\textbf{Ablation study.}
In the Tab.~\ref{tab:ablation}, we make an ablation study to justify the effectiveness of each component. We use the structure of the KPN(K=7) as the `plain' model. We append post-refinement on it(i.e., `w/ PR') and obtain 0.37 dB gain. Then, we analyze the effectiveness of CDM and MDA by incorporating `w/ CDM' and `w/ MDA'. As Tab.~\ref{tab:ablation} show, the CDM bring 0.48 dB promotion and the MDA give 0.09 gain. Since the `w/ CDM' achieves competitive result yet, the MDA component still brings significant improvement. 

\textbf{Efficiency.}
As illustrated in~\ref{tab:efficiency}, our model surpasses traditional image SR(e.g., SRResNet, EDSR) with clear efficiency superiority and performance improvement. Compared with RealSR approaches, our model shows comparable efficiency with significant performance improvements. For instance, our model surpasses KPN(K=19) with 0.51 dB gain and 29$\%$ efficiency drop. Compared with LP-KPL, our model achieves significant improvement over performance with higher efficiency.

\section{Conclusion}
In this article, we propose a dual-path dynamic enhancement network for real image SR. To correct and relax pixel displacement, we incorporate a lightweight model for detail manipulation. To capture multi-scale information, we use multiple dynamic kernels with various sizes for information aggregation. Experimental results well justify the effectiveness of the proposed dual-path model.
\ifCLASSOPTIONcaptionsoff
  \newpage
\fi



%\bibliographystyle{IEEEtran}

\bibliographystyle{IEEEtran}
\bibliography{egbib}

\end{document}